\documentclass[10pt,letterpaper]{article}
\usepackage[top=0.85in,left=1in,footskip=0.75in]{geometry}

\usepackage{amsmath,amssymb}

\usepackage{changepage}

\usepackage[utf8x]{inputenc}

\usepackage{textcomp,marvosym}

\usepackage{cite}

\usepackage{nameref,hyperref}

\usepackage{microtype}
\DisableLigatures[f]{encoding = *, family = * }

\usepackage[table]{xcolor}

\usepackage{array}

\usepackage{multirow}

\newcolumntype{+}{!{\vrule width 2pt}}

\newlength\savedwidth

\newcommand\thickhline{\noalign{\global\savedwidth\arrayrulewidth\global\arrayrulewidth 2pt}%
\hline
\noalign{\global\arrayrulewidth\savedwidth}}

\raggedright
\usepackage[aboveskip=1pt,labelfont=bf,labelsep=period,justification=raggedright,singlelinecheck=off]{caption}

\makeatletter
\renewcommand{\@biblabel}[1]{\quad#1.}
\makeatother

\usepackage[normalem]{ulem}

\usepackage{lastpage,fancyhdr,graphicx}
\usepackage{epstopdf}
\fancyheadoffset[L]{2.25in}
\fancyfootoffset[L]{2.25in}
\begin{document}
\vspace*{0.2in}

\begin{flushleft}
{\Large
\textbf\newline{TweepFake: about detecting deepfake tweets} % Please use "sentence case" for title and headings (capitalize only the first word in a title (or heading), the first word in a subtitle (or subheading), and any proper nouns).
}
\newline
\\
Tiziano Fagni\textsuperscript{1},
Fabrizio Falchi\textsuperscript{2*},
Margherita Gambini\textsuperscript{3}
Antonio Martella\textsuperscript{4},
Maurizio Tesconi\textsuperscript{1}
\\
\bigskip
\textbf{1} Istituto di Informatica e Telematica - CNR, Pisa, Italy
\\
\textbf{2} Istituto di Scienza e Tecnologie dell'Informazione ``A. Faedo'' - CNR, Pisa, Italy
\\
\textbf{3} University of Pisa, Italy
\\
\textbf{4} University of Trento, Italy
\\
\bigskip

% Insert additional author notes using the symbols described below. Insert symbol callouts after author names as necessary.
% 
% Remove or comment out the author notes below if they aren't used.
%
% Primary Equal Contribution Note
%\Yinyang These authors contributed equally to this work.

% Additional Equal Contribution Note
% Also use this double-dagger symbol for special authorship notes, such as senior authorship.
%\ddag These authors also contributed equally to this work.
The authors contributed equally to this work.

%\textcurrency Current Address: Dept/Program/Center, Institution Name, City, State, Country % change symbol to "\textcurrency a" if more than one current address note
% \textcurrency b Insert second current address 
% \textcurrency c Insert third current address


% Group/Consortium Author Note
%\textpilcrow Membership list can be found in the Acknowledgments section.

% Use the asterisk to denote corresponding authorship and provide email address in note below.
* fabrizio.falchi@cnr.it

\end{flushleft}
\section*{Abstract}
The recent advances in language modeling significantly improved the generative capabilities of deep neural models: in 2019 OpenAI released GPT-2, a pre-trained language model that can autonomously generate coherent, non-trivial and human-like text samples. Since then, ever more powerful text generative models have been developed. Adversaries can exploit these tremendous generative capabilities to enhance social bots that will have the ability to write plausible \textit{deepfake} messages, hoping to contaminate public debate. To prevent this, it is crucial to develop deepfake social media messages detection systems. However, to the best of our knowledge no one has ever addressed the detection of machine-generated texts on social networks like Twitter or Facebook. With the aim of helping the research in this detection field, we collected the first dataset of \emph{real} deepfake tweets, \textit{TweepFake}. It is \emph{real} in the sense that each deepfake tweet was actually posted on Twitter. We collected tweets from a total of 23 bots, imitating 17 human accounts. The bots are based on various generation techniques, i.e., Markov Chains, RNN, RNN+Markov, LSTM, GPT-2. We also randomly selected tweets from the humans imitated by the bots to have an overall balanced dataset of 25,572 tweets (half human and half bots generated). The dataset is publicly available on Kaggle. Lastly, we evaluated 13 deepfake text detection methods (based on various state-of-the-art approaches) to both demonstrate the challenges that Tweepfake poses and create a solid baseline of detection techniques. We hope that TweepFake can offer the opportunity to tackle the deepfake detection on social media messages as well.
%We hope that the scale, accuracy, diversity and hierarchical structure of ImageNet can offer unparalleled opportunities to researchers in the computer vision community and beyond.

% Please keep the Author Summary between 150 and 200 words
% Use first person. PLOS ONE authors please skip this step. 
% Author Summary not valid for PLOS ONE submissions.   
%\section*{Author summary}

% TO DO FOR SUBMISSION!!!!!!!!
%\linenumbers

% Use "Eq" instead of "Equation" for equation citations.

\section*{Introduction}

During the last decade, the social media platforms - developed to connect people and make them share their ideas and opinions through multimedia contents (like images, video, audio, and texts) - have also been used to manipulate and alter the public opinion thanks to \href{https://en.wikipedia.org/wiki/Internet_bot}{\textit{bots}}, i.e., computer programs that control a fake social media account as a legitimate human user would do: by ``liking'', sharing and posting old or new media which could be real, forged through simple techniques (e.g., editing of a video, use of gap-filling texts and search-and-replace methods) or deepfake.

Deep learning is a family of machine-learning methods that use artificial neural networks to learn a hierarchy of representations, from low to high non-linear features representation, of the input data. The term \emph{deepfake} is a portmanteau of \emph{deep learning} and \emph{fake}; it refers to AI-generated multimedia (images, videos, audios, and texts) that are potentially deceptive \cite{vincent_2018_deepfake_meanining}, although good usages of deepfakes can be found \cite{youtube_2019_dali_lives}.
The generation and sharing of deepfake multimedia over social media, tricking people into believing that they are human-generated, have already caused distress in several fields (such as politics and science). Therefore, it is necessary to continuously probe the generative model's ability to produce deceiving multimedia by enhancing and developing appropriate detectors. This is more than ever necessary for the text generation field: in 2019, for the first time, a generative model (GPT-2 language model \cite{radford2019gpt2}) showed incredible text generation capabilities that deeply worries the research community: \cite{gehrmann2019gltr} and \cite{adelani2019generating} proved that humans seem unable to identify automatically generated text (their accuracy is near random guessing, i.e. 54\%). Deepfake social media texts (GPT-2 samples included) can already be found, though there is still no misuse episode on them.

Deepfake detecting strategies are continuously developed, from deepfake video  \cite{tolosana2020deepfakes, lyu2020deepfake, nguyen_2019_deep} to audio \cite{chen2020generalization} and text detection methods.
%In \cite{li2018ictu, Deepfake_detection_never_enough_2019_online}, Li stated that because deepfake videos are usually generated with a GAN approach, with the subsequent detectors exploiting flaws in deepfake generation, the detectors are just replicating the GAN dynamic in the wider research landscape, where each new deepfake detection paper gives the deepfake makers a new challenge to overcome, as it was in the eye blinking case. This could result in an endless run of deepfake video generation and detection, as the endless fight between computer viruses and anti-viruses. However, the conflict between generators and detectors could find an end in the deepfake text field, as current machine-text generation techniques exploit language modeling rather than the GAN approach. Anyway, the deepfake text battle is still ongoing too.
%- even though Delip Rao - VP of research at the AI Foundation – noted that even deepfake detectors with an accuracy around \(97\%\) cannot be enough when thinking at the scale of internet platforms \cite{Deepfake_detection_never_enough_2019_online} which deal with millions of multimedia contents every day.
Under the hood, current automatic neural text detectors tend to learn not to discriminate between neural text and human-written text, but rather decide what is characteristic and uncharacteristic of neural text~\cite{Wolff2020AttackingNT} (i.e., statistics of the language for machine-generated texts); however, it emerged that some strategies (substituting homoglyphs to characters or adding some common misspelled words) can alter the statistical characteristics of the generated text making the detection task ever more difficult~\cite{Wolff2020AttackingNT}. Moreover, nowadays scientific works focus on \textit{ad-hoc} generated texts only; also, deep fake detectors usually run knowing the adversarial generative model. This is a white-box approach; \cite{bakhtin2019real_fake} studied the black-box approach (pretending not knowing the text generator), but the text samples were always \textit{ad-hoc} generated. Besides, the majority of the studies deal with long deepfake texts like news articles or stories: according to \cite{ippolito2019automatic_detection_generated_text}, ``For both human raters and automatic discriminators, the longer the provided text excerpt is, the more easily its provenance can be identified''. Having said that, there is a lack of knowledge on how state-of-the-art detection techniques perform in a \textit{real social media setting}, in which the machine-generated text samples are the ones actually posted on a social media, the social media content is often short (above all on Twitter) and the generative model is not known (also, the text samples can be altered to make difficult the automatic detection).

Additionally, to the best of our knowledge, a properly labeled dataset containing \textit{only} human-written and \textit{real} deepfake social media messages still doesn't exist. \cite{garcia-silva-etal-2019-empirical} and \cite{lundberg2018fly} tried to detect auto-generated tweets over a dataset of tweets produced by a large variety of bots \cite{gorwa2018unpacking} (spam bots, social bots, sockpuppet, cyborgs), meaning that the detection task is \textit{not} focused \textit{only} on \textit{real} deepfake messages. Furthermore, those tweets are \textit{human}-labelled at the account level, i.e., by examining the messages produced by a user, but \cite{cresci2017paradigm} proved that a human is not reliable on this labeling task. 

Our work provides the \textit{first} properly labeled dataset of human and real machine-generated social media messages (coming from Twitter in particular): \emph{TweepFake - A Twitter Deep Fake Dataset}. TweepFake contains \emph{real} deepfake tweets that we collected with the goal of testing existing and future detection approaches. The dataset is \emph{real} in the sense that each deepfake tweet was actually posted on Twitter. We collected tweets from a total of 23 bots, imitating 17 human accounts. The bot accounts are based on various generation techniques, including Markov Chains, RNN, RNN+Markov, LSTM, GPT-2. We randomly selected tweets from the humans imitated by the bots to have an overall balanced dataset of 25,572 tweets (half human and half bots generated). We made TweepFake publicly available on Kaggle \cite{TweepFake_2020_online}. More information can be found in the Subsection \emph{The TweepFake dataset}.
With the aim of showing the challenges that TweepFake poses and providing a solid baseline of detection techniques, we also evaluated 13 different deepfake text detection methods: some of them exploiting text representations as inputs to machine-learning classifiers, others based on deep learning networks, and others relying on the fine-tuning of transformer-based classifiers. 
The code used in the experiments is publicly available on GitHub \cite{scripts}.

\subsection*{Related work}

Deepfake technologies have first risen in the computer vision field  \cite{bregler1997video, Suwajanakorn_2017_synthesizing_obama, thies_2016_face2face, chan2019everybody}, followed by effective attempts on audio manipulation \cite{FTC_2020_audio_df, jia2018transfer} and text generation \cite{radford2019gpt2}. Deepfakes in computer vision usually deal with face manipulation - such as entire face synthesis, identity swap, attribute manipulation, and expression swap \cite{tolosana2020deepfakes} – and body re-enacting \cite{chan2019everybody}. Recently, audio deepfakes involved the generation of speech audio from a text corpus by using the voice of different speakers after five seconds of listening time \cite{jia2018transfer}. In 2017, the development of the \textit{self-attention} mechanism and the \cite{vaswani2017attention}'s transformer led to the improvement of the language models. Language modeling refers to the use of various statistical and probabilistic techniques to determine the probability of a given sequence of words occurring in a sentence. The subsequent transformer-based language models (GPT \cite{radford2018gpt}, BERT \cite{devlin2018bert}, GPT2 \cite{radford2019gpt2} etc.) did not only enhance the natural language understanding tasks, but language generation as well. In 2019, \cite{radford2019gpt2} developed GPT-2, a pre-trained language model that can \textit{autonomously} generate coherent human-like paragraphs of text by having in input just a short sentence; in the same year, \cite{zellers2019grover} contributed to text generation with GROVER, a new approach for efficient and effective learning and generation of multi-field documents such as journal articles. Soon after, \cite{keskar2019ctrl} released CTRL, a conditional language model that uses control codes to generate text having a specific style, content, and task-specific behavior. Last but not least, \cite{li2020optimus} presented OPTIMUS , putting the variational autoencoder in the game for text generation.

Currently, the approaches to automatic deepfake text detection roughly fall into three categories, here listed in order of complexity:
\begin{itemize}
\item[] Simple classifier: a machine-learning or deep-learning binary classifier trained from scratch.
\item[] Zero-shot detection: using the output from a pre-trained language model as features for the subsequent classifier. This classifier may be a machine-learning based one or a simple feed forward neural network.
\item[] Fine-tuning based detection: jointly fine-tuning a pre-trained language model with a final simple neural network (consisting of one or two layers).
\end{itemize}

The GPT-2's research group made an in-house detection research \cite{solaiman2019release} on GPT-2 generated text samples: first, they evaluated a standard machine-learning approach that trains a logistic regression discriminator on \textit{tf-idf} unigram and bigram features. Then, they tested a simple zero-shot baseline using a threshold on the \textit{total probability}: a text excerpt is predicted as machine-generated if its likelihood according to GPT-2 is closer to the mean likelihood over all machine-generated texts than to the mean of the human-written ones.    

\cite{gehrmann2019gltr} provided GLTR (Giant Language model Test Room), a visual tool that helps humans to detect deepfake texts. Generated text is sampled word by word from a next token distribution (several sampling techniques can be used \cite{gen_sampling_tech_huggingface_2020}, the simplest way is to take the most probable token): this distribution usually differs from the one that humans subconsciously use when they write or speak. GLTR tries to show these statistical language differences to aid people in discriminating human-written text samples from machine-generated ones.

GROVER's authors \cite{zellers2019grover} followed the fine-tuning based detection approach by using BERT, GPT2 and GROVER itself as the pre-trained language model. GROVER was the best, suggesting that maybe the best defense against the transformer-based text generators is a detector based on the same kind of architecture. However, OpenAI \cite{solaiman2019release} proved it wrong on GPT-2 generated texts: they showed that fine-tuning a RoBERTa-based detector achieved consistently higher accuracy than fine-tuning a GPT-2 based detector with equivalent capacity.

\cite{bakhtin2019real_fake} developed an energy-based deepfake text detector: unlike auto-regressive language models (e.g. GPT-2 \cite{radford2019gpt2}, XLNET \cite{yang2019xlnet}), which are defined in terms of a sequence of conditional distributions, an energy-based model is defined in terms of single scalar energy function, representing the joint compatibility between all input variables. Thus, the deepfake discriminator is an energy function that scores the joint compatibility of an input sequence of tokens given some context (e.g. a text sample, some keywords, a bag of words, a title) and a set of network parameters. These authors tried also to generalize the experimental setting, where the generator architectures and the text corpora are different between training and test time.

The only research on the detection of deepfake social media texts was conducted by \cite{adelani2019generating} on Amazon reviews written by GPT-2. They evaluated several human-machine discriminators: the Grover-based detector, GLTR, RoBERTa-based detector from OpenAI and a simple ensemble that fused these detectors using logistic regression at the score level. 

The above deepfake text detection methods have got two flaws: except for \cite{adelani2019generating}'s research, they dealt with generated news articles, having a longer length with respect to social media messages; then, just a single known adversarial generative model is usually used to generate the deepfake text samples (usually GPT-2 or GROVER). In a real-setting scenario, we don't know how many and what generative architectures are used. Our Tweepfake dataset provides a set of tweets produced by several generative models, hoping to help the research community in detecting shorter deepfake texts written by heterogeneous generative techniques.

\section*{DeepFake tweets generation}
%\section*{DeepFake Tweets}
There exist several methods to generate a text. What follows is a short description of the generative methods used to produce the machine-generated tweets contained in our dataset.

\subsection*{Generation techniques}
%\subsection*{Methods}
First and foremost, the training set of text corpora is tokenized (punctuation included), and then one of the following methods can be applied. Notice that the following techniques write a text token-by-token (a token could be a word, a char, a byte pair, a Unicode code point) until a stop token is encountered or a pre-defined maximum length is reached. \textit{RNN}, \textit{LSTM}, \textit{GPT2} are language models. Therefore, at each token generation, they always produce a multinomial distribution - in which a category is a token of the vocabulary derived from a set of human-written texts - from which the next token is sampled with a specific sampling technique (e.g., max probability, top-k, nucleus sampling \cite{sampling_techniques}). A special start token is given in input to the generative model to prime the text generation; with language models, a short sentence can work as a priming text as well: each token of the start sentence is processed without computing the multinomial distribution, just to \textit{condition} the generative model.
\begin{itemize}

\item[] \emph{Markov Chains} is a stochastic model that describes a sequence of states by moving from a state to another with a probability which depends on the current state only. For the text generation a state is identified as a token: the next token/state is randomly selected from a list of tokens following the current one. The probability of a token \textit{t} to be chosen is proportional to the frequency of the appearance of \textit{t} after the current token.
\item[] \emph{RNN}, helped by its loop structure, stores in its \textit{accumulated} memory the information on the previously encountered tokens and computes the multinomial distribution from which the next token is chosen. The selected token is given back in input so that the RNN can produce the following one.
\item[] \emph{RNN+Markov} method \textit{may} employ the Markov Chain's next token selection as a sampling technique. In practice, the next token is randomly sampled from the multinomial distribution produced by RNN, with the tokens having the highest probability value being the most likely to be chosen. However, no reference was found to confirm our hypothesis on RNN+Markov mechanism.

\item[] \emph{LSTM} generates text as RNN does. However, it is smarter than the latter because of its more complicated structure: it can learn to selectively keep track of only the relevant information of the already seen piece of text \textit{while} also minimizing the vanishing gradient problem that affects a RNN. LSTM's memory is ``longer'' than RNN's.

\item[] \emph{GPT-2} is a generative pre-trained transformer language model relying on the \textit{Attention} mechanism of \cite{vaswani2017attention}: by employing the \textit{Attention}, a language model pre-trained on millions of sentences/texts learns how each token/word relates to every other in every possible context. This is the trick to generate more coherent and non-trivial paragraphs of text. Anyhow, being a language model, GPT-2's text generation steps are the same as RNN and LSTM: generation of a multinomial distribution at each step and then selection of the next token from it by using a specific sampling technique.

\item[] \emph{CharRNN} employs RNN at c\textit{har level} to generate a text char-by-char.

\end{itemize}

\subsection*{The TweepFake dataset}
In this section we describe the process of building the novel \emph{TweepFake - A Twitter Deep Fake Dataset} together with the results of the experimentation on the deepfake detection task.
Twitter accounts have been searched heuristically on the web, GitHub and Twitter looking for keywords related to automatic or AI text generation, deepfake text/tweets, or to specific technologies as well as GPT-2, RNN, etc. in order to collect a sample of Twitter profiles as huge as possible.
We selected only accounts referring to automated generated text technologies in Twitter descriptions, profile URLs, or related GitHub. From this sample, we selected a subset of accounts mimicking (often fine-tuned on) human Twitter profiles. Thus, we obtained 23 bots and 17 human accounts because some fake accounts imitate the same human profile (see Table \ref{Table:TweetsByAccount}).
Then we downloaded timelines of both deep fake accounts and their corresponding humans via Twitter REST API.  
In order to get a data set balanced on both categories (human and bots) we randomly sampled tweets for each accounts' couple (human and bot/s) based on the less productive. For example, after the download, we had 3,193 tweets by human\#11 and 1,245 by the corresponding bot\#12, thus we random sampled 1,245 tweets by the human account timeline to get the same amount of data. In total, we had 25,572 tweets half human and half bots generated.
In Table \ref{Table:TweetsByAccount}, we report, for each fake account we considered, the human account imitated, the technology used for generating the tweets, and the number of tweets we collected from both the fake and the human account.
In Table \ref{Table:TweetsByTech}, we grouped the fake accounts by technology reporting, together with the number of collected tweets, the citation of the information we found about the bot (i.e., more technical information, code, news, etc.).
Please note that in our detection experiments we grouped the technologies in three main groups: GPT-2, RNN, others (see Sections \emph{Results} and \emph{Discussion}).

% Place tables after the first paragraph in which they are cited.
\begin{table}[hbt!]
\begin{adjustwidth}{0in}{0in} % Comment out/remove adjustwidth %environment if table fits in text column.
%\begin{adjustwidth}{-2.25in}{0in} % Comment out/remove adjustwidth %environment if table fits in text column.
\centering
\caption{
{\bf The proposed \emph{TweepFake} dataset tweets grouped by imitated human account.}
\label{Table:TweetsByAccount}
}
\begin{tabular}{|l|r|l+l|r|}
\hline
\multicolumn{1}{|l|}{\bf fake account} & \multicolumn{1}{|l|}{\bf tweets} & \multicolumn{1}{|l+}{\bf technology} & \multicolumn{1}{l|}{\bf human account } &
\multicolumn{1}{|l|}{\bf tweets}\\ \thickhline
	bot\#1 & 946 & RNN + Markov & human\#1 & 946 \\ \hline
	bot\#2 & 348 & GPT-2 & human\#2 & 348 \\ \hline
	bot\#3 & 132 & GPT-2 & human\#3 & 132 \\ \hline
	bot\#4 & 1792 & GPT-2 & \multirow{2}{*}{human\#4} & \multirow{2}{*}{1803} \\
	bot\#5 & 11 & RNN & &  \\ \hline
	bot\#6 & 38 & LSTM & \multirow{2}{*}{human\#5} & \multirow{2}{*}{56}  \\
	bot\#7 & 18 & Torch RNN & &  \\ \hline
	bot\#8 & 217 & GPT-2 & human\#6 & 217 \\ \hline
	bot\#9 & 1289 & RNN + Markov & human\#7 & 1289 \\ \hline
	\multirow{2}{*}{bot\#10} & \multirow{2}{*}{1030} & \multirow{2}{*}{Markov Chains} & human\#8 & 515 \\
	&  &  & human\#9 & 515 \\ \hline
	bot\#11 & 2409 & RNN & human\#10 & 2409 \\ \hline
	bot\#12 & 1245 & RNN & human\#11 & 1245 \\ \hline
	bot\#13 & 228 & GPT-2 & human\#12 & 228 \\ \hline
	bot\#14 & 355 & GPT-2 & \multirow{6}{*}{human\#13}  & \multirow{6}{*}{1293}  \\ 
	bot\#15 & 33 & GPT-2 &   &  \\ 
	bot\#16 & 286 & RNN &  &  \\ 
	bot\#17 & 549 & GPT-2 &  &  \\
	bot\#18 & 18 & GPT-2 &  &  \\ 
	bot\#19 & 52 & unknown &  &  \\ \hline
	bot\#20 & 100 & CharRNN & human\#14 & 100 \\ \hline
	bot\#21 & 39 & GPT-2 & human\#15 & 39 \\ \hline
	bot\#22 & 128 & OpenAI & human\#16 & 128 \\ \hline
	bot\#23 & 1523 & unknown & human\#17 & 1523 \\ \hline

\end{tabular}
% \begin{flushleft} Table notes Phasellus venenatis, tortor nec vestibulum mattis, massa tortor interdum felis, nec pellentesque metus tortor nec nisl. Ut ornare mauris tellus, vel dapibus arcu suscipit sed.
% \end{flushleft}
% \label{table1}
\end{adjustwidth}
\end{table}

% Place tables after the first paragraph in which they are cited.
\begin{table}[hbt!]
%\begin{adjustwidth}{-2.25in}{0in} % Comment out/remove adjustwidth %environment if table fits in text column.
\centering
\caption{
{\bf The proposed \emph{TweepFake} dataset tweets grouped by technology.}
\label{Table:TweetsByTech}
}
\begin{tabular}{|c+l|l|c|c|r|r|}
\hline
\multicolumn{1}{|l+}{\bf technology} &
\multicolumn{1}{|l|}{\bf fake acc.} &
\multicolumn{1}{l|}{\bf human acc. } &
\multicolumn{1}{l|}{\bf info } &
\multicolumn{1}{|l|}{\bf code } &
\multicolumn{1}{l|}{\bf tweets} &
\multicolumn{1}{|l|}{\bf tweets} \\ \thickhline

%GPT-2
\multirow{10}{*}{GPT-2} & bot\#2 & human\#2 & \cite{radford2019gpt2} & \cite{whalefakesCode} & 348 & \multirow{10}{*}{3711} \\ 
& bot\#3 & human\#3 & \cite{calebgamman2} & \cite{calebgamman2Code} & 132 &  \\ 
& bot\#4 & human\#4 & - & \cite{drilgpt2Code}& 1792 &  \\ 
& bot\#8 & human\#6 & - & \cite{GenePark_GPT2Code} & 217 &  \\ 
& bot\#13 & human\#12 & 228 & - & - &   \\ 
& bot\#14 & human\#13  &  - & \cite{botustrumpCode} & 355 &   \\ 
& bot\#15 & human\#13 & - & - & 33 &  \\  
& bot\#17 & human\#13 & - & - & 549 &   \\ 
& bot\#18 & human\#13 & - & - & 18 &   \\ 
& bot\#21 & human\#15 & - & - & 39 &  \\ \hline

% RNN
\small\multirow{5}{*}{RNN} & bot\#5 & human\#4	  & - & - & 11 & \multirow{5}{*}{3969} \\
  & bot\#11 & human\#10  &\cite{Hooke}& - &2409 &  \\
  & bot\#12 & human\#11  & \cite{AI-NarendraModi} &  \cite{AI-NarendraModiCode} & 1245 &  \\
  & bot\#16 & human\#13  & \cite{DeepDrumpf} &  \cite{DeepDrumpfCode} & 286 & \\ 
 & bot\#7 & human\#5   & \cite{Musk_from_Mars} & \cite{Musk_from_MarsCode} & 18 & \\ \hline

% RNN + Markov
\multirow{2}{*}{RNN + Markov}  & bot\#1 & human\#1 & - & - & 946  & \multirow{2}{*}{2235}\\ 
& bot\#9 & human\#7 & - & - & 1289 &  \\ \hline

% MARKOV CHAINS
\multirow{2}{*}{Markov Chains} &
\small\multirow{2}{*}{bot\#10} & \small human\#8   &
\small\multirow{2}{*}{\cite{theJadenTrudeau}} & \small\multirow{2}{*}{\cite{theJadenTrudeauCode}} & \multirow{2}{*}{1030} & \multirow{2}{*}{1030} \\
	 & & \small human\#9 & &  &  & \\ \hline

LSTM & bot\#6 & human\#5 & \cite{DeepElonMusk} & \cite{DeepElonMuskCode} & 38 & 38   \\ \hline
	
OpenAI & bot\#22 & human\#16 & - & - & 128 & 128  \\ \hline
CharRNN & bot\#20 & human\#14 & - & \cite{deep_thorinCode}& 100 & 100 \\ \hline
unknown & bot\#19 & human\#13 & - & - & 52 & 52  \\ \hline
unknown & bot\#23 & human\#17 & - & - & 1523 & 1523  \\ \hline

\end{tabular}
% \begin{flushleft} Table notes Phasellus venenatis, tortor nec vestibulum mattis, massa tortor interdum felis, nec pellentesque metus tortor nec nisl. Ut ornare mauris tellus, vel dapibus arcu suscipit sed.
% \end{flushleft}
% \label{table1}
%\end{adjustwidth}
\end{table}

\section*{DeepFake tweets detection}

\subsection*{Detection techniques}

To verify the difficulty level in the detection task of automatically generated natural language contents, we used the built dataset to measure the effectiveness of a set of ML and DL methods of increasing complexity. The results obtained allow us to fix some baseline configurations in terms of performance and give an idea on which approaches are most promising in solving this specific problem.

In Table \ref{tab:methods_description}, we report all the methods that have been tested in this work. We explored the usage of four main approaches to model the solutions to this specific task. The first scenario uses a text representation based on bag-of-words (BoW)\cite{sebastiani_textcat} with encoded feature weighted according to TF-IDF function\cite{sebastiani_textcat}. The tweets encoded in this way have been next processed by a statistical ML algorithm able to produce a suitable classifier to solve the specific problem. In this work, we have chosen to implement three popular classifiers:  logistic regression, random forest, and SVM.

\begin{table}[hbt!]
%\begin{adjustwidth}{-2.25in}{0in} % Comment out/remove adjustwidth environment if table fits in text column.
		\centering
		\caption{
			{\bf Description of the methods used in the experimentation.}}
		\begin{tabular}{|c|l|l|}
			\hline
			\textbf{Encoding} & \textbf{Method name} & \textbf{Algorithm}\\
			\hline
			
	    	\multirow{3}{*}{BoW+TF\_IDF} & \textsc{log\_reg\_bow} &  Logistic regression classifier\cite{nigam1999using} \\
			
			& \textsc{rand\_forest\_bow} & Random forest classifier\cite{breiman2001random} \\

		    & \textsc{svc\_bow}  & Support vector machine classifier\cite{joachims1998text}\\

			\hline
			
			\multirow{3}{*}{BERT} & \textsc{log\_reg\_bert} & Logistic regression classifier\\

			& \textsc{rand\_forest\_bert} & Random forest classifier\\

			& \textsc{svc\_bert} & Support vector machine classifer\\

			\hline
			
			\multirow{6}{*}{Characters} & \textsc{char\_cnn}  & Single CNN network\cite{lecun1998gradient} using internal char \\
			& & embeddings representation\\

			& \textsc{char\_gru} & Single GRU network\cite{cho-etal-2014-learning} using internal char\\
			& & embeddings representation\\

			& \textsc{char\_cnngru} & Combined CNN and GRU networks using\\
			& & internal char embeddings representation\\

			\hline
			\multirow{4}{*}{Native LM } & \textsc{bert\_ft} & BERT language model with fine-tuning\\

			& \textsc{distilbert\_ft} & DistilBERT language model with fine-tuning\\

			& \textsc{roberta\_ft} & RoBERTa language model with fine-tuning\\

			& \textsc{xlnet\_ft} & XLNet language model with fine-tuning\\
			\hline
		\end{tabular}
		%\begin{flushleft} Table notes Phasellus venenatis, tortor nec vestibulum mattis, massa tortor interdum felis, nec pellentesque metus tortor nec nisl. Ut ornare mauris tellus, vel dapibus arcu suscipit sed.
		%\end{flushleft}
		\label{tab:methods_description}
%    \end{adjustwidth}
\end{table}

The approach based on BoW+TF-IDF, although being very popular and used for many years as the primary methodology to vectorize texts, suffers from two main drawbacks. The first problem is related to the curse of dimensionality\cite{verleysen2005curse}, i.e., the feature space is very sparse, and the amount of data required to produce statistically significant models is very high. The second issue of BoW is that it ignores the information about word order, and thus it misses completely any information about the semantic context on which the words occur. To overcome these limitations, on the second approach, we encoded texts using BERT\cite{devlin2018bert}, a recent pre-trained language model that contributed to improving state-of-the-art results on many  NLP problems. BERT provides contextual embeddings, fixed-size vector representations of words which depend not only by the words itself but also by the context on which the words occur: for example, the word \emph{bank}, depending on being near to word  \emph{economy} or \emph{river}, will assume a different meaning and consequently a different contextual vector. Therefore, these contextual representations can be merged together to obtain a contextualized fixed-size vector of a specific text (e.g., by averaging the vectors of all words composing a specific text). As in the previous scenario, the tweets encoded through BERT has been processed using the same set of classifiers.

\begin{figure}[!h]
%\begin{adjustwidth}{-2.25in}{0in}
\includegraphics[width=\textwidth]{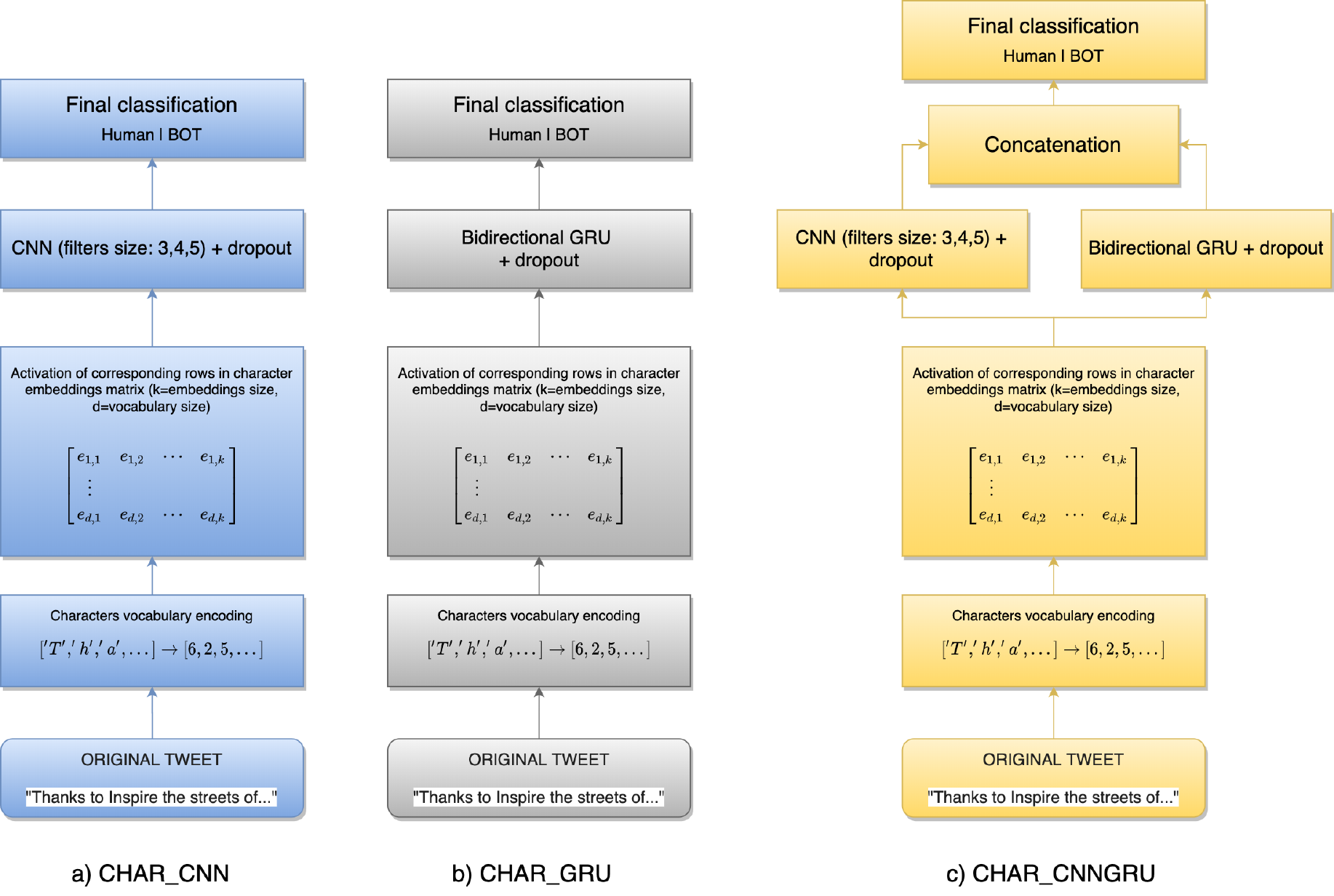}
\caption{{\bf Architecture of the tested deep neural networks based on character encoding.}
Three different architectures were tested: a) a CNN sub-network using three different kernel sizes (3,4, and 5) combined together and followed by a dropout layer, b) a bidirectional GRU followed by a dropout layer, and c) a network exploiting both CNN and GRU to extract spatial and temporal features from data in order to try to improve the effectiveness of the solution.
}
\label{Fig 1}
%CharDLArchitecture
%\end{adjustwidth}
\end{figure}

On the third approach, we leverage another effective way to encode textual contents by working at the character level instead of words or tokens\cite{NIPS2015_5782}. This methodology has the advantage of not requiring access to any external resource, but it only exploits the dataset used to learn the model. The encoding process is summarized in Fig~\ref{Fig 1}. 

Each tweet is encoded as a set of contiguous characters IDs obtained from a fixed vocabulary of characters. This mapping allows us to use the internal embeddings matrix (learned during training phase) to select, at each time step in the text, only the row vector corresponding to the current analyzed character, thus contributing to building a proper matrix representation of current text. The resulting text's embedding matrix is next passed as input to the successive layers in the tested deep learning networks.

As the final and most effective approach, we used several pre-trained language models by fine-tuning them directly on the built dataset. This process just consists of taking a trained language model, integrate its original architecture with a final dense classification layer, and perform training on a specific dataset (typically small) for very few epochs\cite{devlin2018bert}. This step of fine-tuning allows us to customize and adapt the native language model's network weights to the considered use case, maximizing the encoding power of the model to solve that specific classification problem. As reported in Table \ref{tab:methods_description}, in this work, we tested four different language models, all based on transformer architecture\cite{vaswani2017attention}, which have provided state of the art results on many text processing benchmarks. BERT\cite{devlin2018bert} was presented in 2018, and thanks to the innovative transformer-based architecture with dual prediction tasks (Masked Language Model and Next Sentence Prediction) and much data, it was able to basically outperform all other methods on many text processing benchmarks. XLNet\cite{yang2019xlnet} and RoBERTa\cite{liu2019roberta} tried to increase BERT effectiveness by slightly varying and optimizing its original architecture, and using a lot more data on training step, resulting in improvements on prediction powers on the same benchmarks up to 15\%. DistilBERT\cite{liu2019roberta}, on the other hand, tried to keep the performance of the original BERT model (97\% of original ones)  but greatly simplifying the network architecture and halving the number of parameters to be learned. 

\subsection*{Experimental setup}

The main parameters of each algorithm (except for those based on deep learning models where, for computational reasons, we used default parameters) have been optimized using the validation set.

Baselines built on standard machine learning methods (with both BoW and BERT representations) have been implemented using \verb|scikit-learn| Python library \cite{sklearn_api}. In BoW experiments, we performed tweets tokenization by splitting texts into words, removing all hashtags, replacing all user mentions with the token \verb|__user_mention__|, replacing all URLs with token \verb|__url__|, and leaving all found emoticons as separated tokens. During the encoding phase, to minimize computational cost, we only left the most frequent 25,000 tokens, and we weighted each token inside tweets according to tf-idf method\cite{sebastiani_textcat}. In BERT experiments we encoded tweets using \verb|bert-base-cased| pre-trained model from \verb|transformers| Python library\cite{Wolf2019HuggingFacesTS}. In SVC configurations we tried different kernels (\verb|linear| and \verb|rbf|), and a range of values for \verb|C| and \verb|gamma| parameters. The \verb|C| misclassification cost has also been optimized on logistic regression configurations. On random forest baselines we have chosen the best setting varying these parameters: \verb|max_depth|, \verb|min_samples_leaf|, \verb|min_samples_split|, and \verb|n_estimators|. 

Solutions based on characters deep learning networks have been implemented using \verb|Keras| Python library \cite{chollet2015keras}. We used a fixed window of length 280 (on Twitter, 280 is the maximum length of a tweet in terms of the number of characters) to represent input tweets and \verb|tanh| activation function at every level of hidden layers. In all three configurations of chars neural networks, the first hidden layer is an embedding layer of size 32. At the second level, \textsc{char\_cnn} is characterized by three independent CNN subnetworks (CNN layer composed by 128 filters and followed by a global max pooling layer) with different kernel sizes (3,4, and 5) which are next concatenated and filtered by a dropout layer before performing final classification. \textsc{char\_gru} configuration is more simple, composed at the second level by a bidirectional GRU layer followed by dropout and a final classification layer. \textsc{char\_cnngru} configuration adds to the first hidden layer two different subnetworks (one CNN-based and one GRU-based with the same architecture as defined before), concatenates them, applies a dropout, and performs final classification.

We used \verb|simpletransformers| Python library \cite{rajapakse2019simpletransformers} to implement all models in fine-tuned configurations. In agreement with other works in literature and for computational reasons, we decided to limit the number of epochs to just three complete iterations over training data. 

A summary of the customized parameter values used in the final configurations is reported in \nameref{S1_Tab} %Table \ref{tab:optimized_parameters}.
All the other unspecified parameters used by tested algorithms are left to their default values, as defined in the software libraries providing their implementation.

Our experiments are reported in a GitHub repository \cite{scripts}.

%\section*{Results and Discussion}
\section*{Results}
As evaluation measures, we used the canonical adopted metrics in text classification contexts: precision, recall, F1, and accuracy\cite{sebastiani_textcat}. In this context, given that the analyzed dataset is balanced in terms of examples, the accuracy seems the most reasonable measure to capture the effectiveness of a method.
On Table \ref{tab:baseline_results}, we report the results obtained on test set using the proposed detection baselines.

\begin{table}[!ht]
	%\begin{adjustwidth}{-2.25in}{0in} % Comment out/remove adjustwidth environment if table fits in text column.
		\centering
		\caption{
			{\bf Experimental results on test set obtained with the proposed baselines.}}
		\begin{tabular}{|c|ccc|ccc|c|}
			\hline
			\multirow{2}{*}{\bf {Method}} & \multicolumn{3}{c|}{\textsc{HUMAN}} & \multicolumn{3}{c|}{\textsc{BOT}} &  \textsc{GLOBALLY}\\
			 & \bf{Precision} & \bf{Recall} & \bf{F1} & \bf{Precision} & \bf{Recall} & \bf{F1} & \bf{Accuracy}\\ 
			\hline
			\textsc{log\_reg\_bow} & 0.841 & 0.749 & 0.792 & 0.774 & 0.859 & 0.814 & 0.804\\
			\textsc{rand\_forest\_bow} & 0.759 & 0.798 & 0.778 & 0.787 & 0.747 & 0.767 & 0.772\\
			\textsc{svc\_bow} & 0.851 & 0.754 & 0.800 & 0.779 & 0.869 & 0.822 & 0.811\\
			\hline
			\textsc{log\_reg\_bert} & 0.846 & 0.820 & 0.833 & 0.826 & 0.851 & 0.838 & 0.835\\
			\textsc{rand\_forest\_bert} & 0.864 & 0.776 & 0.818 & 0.797 & 0.878 & 0.836 & 0.827\\
			\textsc{svc\_bert} & 0.860 & 0.818 & 0.838 & 0.827 & 0.867 & 0.846 & 0.842\\
			\hline
			\textsc{char\_cnn} & 0.896 & 0.794 & 0.842 & 0.815 & 0.908 & 0.859 & 0.851\\
			\textsc{char\_gru} & 0.899 & 0.743 & 0.814 & 0.781 & 0.916 & 0.844 & 0.830\\
			\textsc{char\_cnngru} & 0.848 & 0.820 & 0.834 & 0.826 & 0.853 & 0.839 & 0.837\\
			\hline
			\textsc{bert\_ft} & 0.899 & 0.882 & 0.890 & 0.884 & 0.901 & 0.892 & 0.891\\
			\textsc{distilbert\_ft} & 0.894 & 0.880 & 0.886 & 0.882 & 0.895 & 0.888 & 0.887\\
			\textsc{roberta\_ft} & 0.901 & \textbf{0.890} & \textbf{0.895} & \textbf{0.891} & 0.902 & \textbf{0.897} & \textbf{0.896} \\
			\textsc{xlnet\_ft} & \textbf{0.914} & 0.832 & 0.871 & 0.846 & \textbf{0.922} & 0.882 &  0.877\\
			\hline
		\end{tabular}
		%\begin{flushleft} Table notes Phasellus venenatis, tortor nec vestibulum mattis, massa tortor interdum felis, nec pellentesque metus tortor nec nisl. Ut ornare mauris tellus, vel dapibus arcu suscipit sed.
		%\end{flushleft}
		\label{tab:baseline_results}
	%\end{adjustwidth}
\end{table}

To have a better understanding on how the tested baselines behave at detection time, we split all available accounts on the dataset into four different categories:
\begin{description}
\item[human] The set of Twitter accounts having contents produced only by a human.
\item[gpt2] The set of Twitter accounts having contents produced only by GPT2-based generative algorithms.
\item[rnn] The set of Twitter accounts having contents produced only by RNN-based generative algorithms.
\item[others] The set of Twitter accounts having contents produced only by generative algorithms using mixed (e.g., RNN + Hidden Markov models)  or unknown approaches.
\end{description}

Each account has been assigned to one of those categories according to the specific information found in the corresponding Twitter account's description or in a linked Web page describing its purpose. For some accounts, we were not able to find any information provided by the author about the technology used to implement the BOT, so in that case we assigned the account to \emph{others} category.

\begin{figure}[!h]
\centerline{\includegraphics[width=\textwidth]{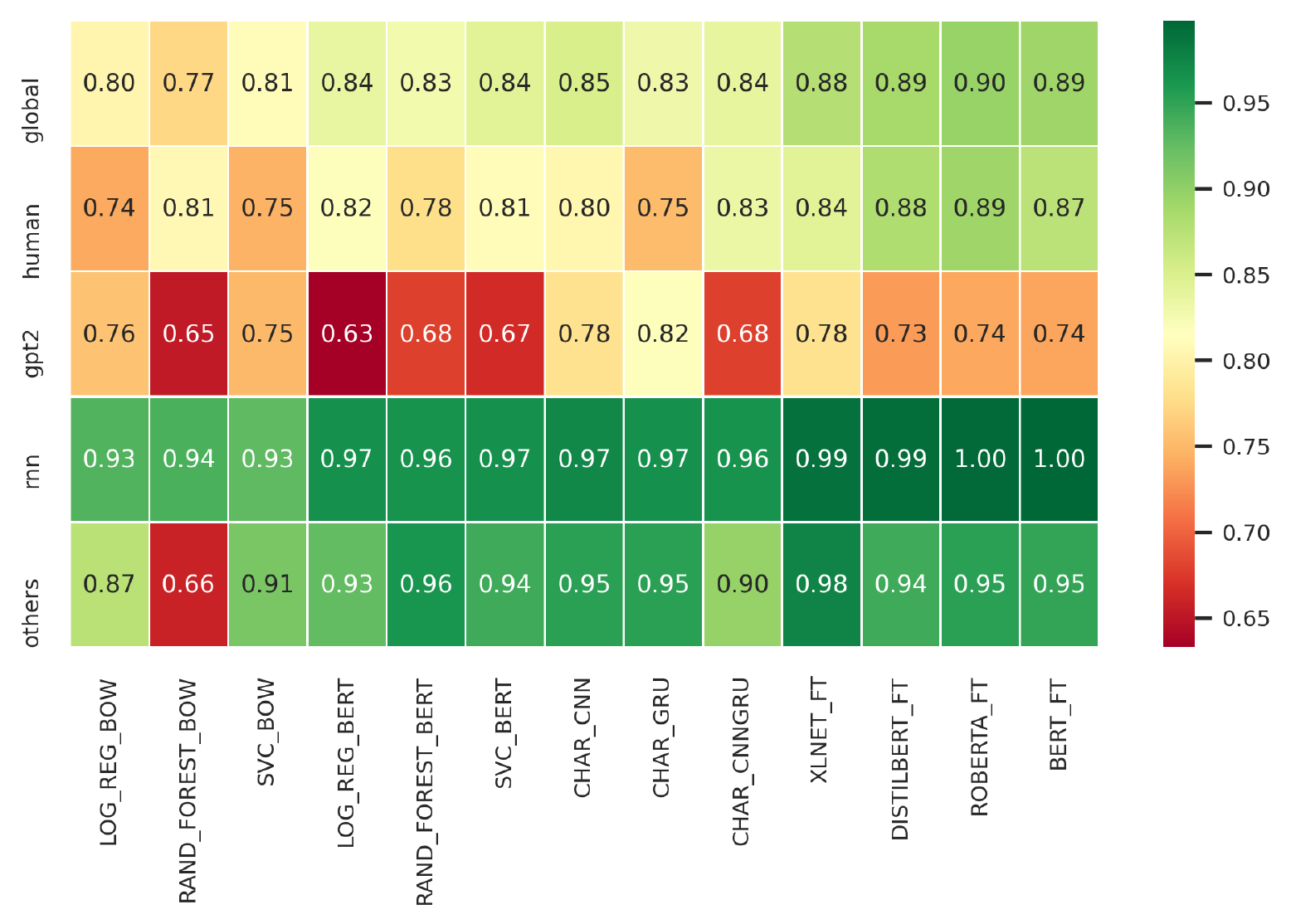}}
\caption{\textbf{Accuracy heat-map over fake account type.}}
\label{Fig 2}
%HeatmapAccuracy
\end{figure}

On Fig~\ref{Fig 2} we show a qualitative evaluation of the accuracy of the proposed baselines in relation to the category of accounts and the ``global'' performance over all categories.

To obtain a fair comparison between \emph{human} and the other types of categories, giving that the \emph{human} class has more examples than the other categories alone, we performed a random undersampling of \emph{humans} to match the maximum size in terms of examples given by one of the other three categories. The resulting distribution in terms of examples has been the following:  \emph{humans} (484), \emph{GPT-2} (384), \emph{RNN} (412), and "others" (484).

\section*{Discussion}

Globally, in terms of accuracy, we can observe (Table \ref{tab:baseline_results}) that the methods based on BoW representations have the worst performance (around 0.80), followed by those using both BERT (around 0.83) and character encodings (up to 0.85), and  remarkably outperformed by methods using native language modeling encoding (0.90 for \textsc{roberta\_ft}). A high level view of the results thus indicates that the most complex text representations, especially those based on big amount of external data, provide evident advantages in terms of effectiveness. An interesting exception is the character encoding in deep learning methods which is simple, able to generally provides good performance, and be useful in cases where no pretrained models are available (e.g., for non-english languages). 

Going more on details, the baselines based on fine tuning (except for XLNET) show very well balanced performance in terms of precision and recall, differently from the other configurations where one the two measures is a lot higher than the other. Another observation is that all methods provide a) higher precision on human label examples than on bot examples ones and b) higher recall on bot label examples than on human ones. This translates into having more difficulties to detect correctly a tweet as written by a bot instead than a human, although the algorithms have more troubles finding all relevant human label examples. 
%\sout{The methods using fine tuning don't have a similar uniform behaviour but in general the differences between precision and recall for both codes are quite small, providing in general competitive accuracy values in all tested conditions.}

The qualitative analysis of the accuracy in relation to the accounts' categories highlights some interesting facts (Fig \ref{Fig 2}): a) all methods (except \textsc{rand\_forest\_bow}) perform extremely well in identifying tweets as BOT on both on \emph{RNN} and \emph{others} accounts; b) tweets from \emph{human} accounts are easily identifiable by methods based on fine tuning but not from the others; and c) all methods have difficulties in identifying correctly  tweets produced by \emph{GPT-2} accounts. In particular, on this last point it is interesting to note that all complex fine tuned LM methods perform remarkably worst than some character based methods like \textsc{CHAR\_GRU}. This could indicate that RNN networks maintain slight advantages in temporal representations for short contexts respect to newer transformer networks, an important aspect to be investigated in the future. 

To sum up, these findings suggest that a wide variety of detectors (text representation-based using machine-learning or deep-learning methods and transformer-based using transfer-learning) have greater difficulties in detecting correctly a deepfake tweet rather than a human-written one; this is especially true for GPT-2 generated tweets, insinuating that the newest and more sophisticated generative methods based on the transformer architecture (here GPT-2) can produce more human-like short texts than old generative methods like RNNs. We manually verified several GPT-2 and RNN tweets: the former were harder to label as bot-generated. In any case, a future work could deeply investigate the humanness of tweets coming from several generative methods by questioning people.

\section*{Conclusion}
Deepfake text detection is increasingly critical due to the development of highly sophisticated generative methods like GPT-2. However, to the best of our knowledge no deepfake detection has been conducted over social media texts yet. Therefore, the aim of this paper was to present the first \emph{real} deepfake tweets dataset (TweepFake) to help the research community to develop techniques fighting the deepfake threat on social media. The proposed \emph{real} deepfake tweets are publicly available on the well-known Kaggle platform. The dataset is composed of 25,572 tweets, half human and half bots generated, posted on Twitter in the last few months. We collected them from 23 bots and from the 17 human accounts they imitate. The bot accounts are based on various generative techniques, including GPT-2, RNN, LSTM and Markov Chain. We tested the difficulty in discriminating human-written tweets from machine-generated ones by evaluating 13 detectors: some of them exploiting text representations as inputs to machine-learning classifiers, others based on deep learning networks, and others relying on the fine-tuning of transformer-based classifiers. 

Our detection results suggest that the newest and more sophisticated generative methods based on the transformer architecture (e.g., GPT-2) can produce high-quality short texts, difficult to unmask also for expert human annotators. This finding is in line with the ones in \cite{radford2019gpt2} covering long texts (news, articles, etc.). Additionally, the transformer-based language models provide very good word representations for both text representation-based and fine-tuning based detection techniques. The latter provide a better accuracy (nearly 90\% for RoBERTa-based detector) than the former. %An interesting exception is the CHAR\_GRU detector, whose specific accuracy on GPT-2 tweets reach 82\%; this may indicate that RNN networks maintain slight advantages in temporal representations for short contexts like tweets.
%Overall, the detection results reported as a baseline using 13 detection methods, confirm that the newest and more sophisticated generative methods based on transformer architecture (e.g., GPT-2) can produce high-quality short texts, difficult to detect.

We recommend to further investigate the RNN-based detectors, as the CHAR\_GRU-based detector was the best at correctly labelling GPT2-tweets as bots. Moreover, a study of the capability of humans to discriminate human-written tweets from machine-generated ones is necessary; also, the humanness of tweets produced by different generative methods could be assessed. Of course, different detection techniques are appreciated.

% \section*{Supporting information}

% % Include only the SI item label in the paragraph heading. Use the \nameref{label} command to cite SI items in the text.

% \paragraph*{S1 Table}
% \label{S1_Tab}
% {\bf Parameter values.} Parameter values used in the final experimentation on the test set.

% \paragraph*{S2 Fig.}
% \label{S2_Fig}
% {\bf Acc. over humans.} Detection Accuracy of tested methods on 'human' accounts with at least 10 examples.

% \paragraph*{S3 Fig.}
% \label{S3_Fig}
% {\bf Acc. over GPT-2.} Detection Accuracy of tested methods on 'gpt2' accounts with at least 10 examples.

% \paragraph*{S4 Fig.}
% \label{S4_Fig}
% {\bf Acc. over RNN.} Detection Accuracy of tested methods on 'rnn' accounts with at least 5 examples.

% \paragraph*{S5 Fig.}
% \label{S5_Fig}
% {\bf Acc. over \emph{others}.} Detection Accuracy of tested methods on 'others' accounts with at least 5 examples.

\section*{Acknowledgments}

This work has been fully supported by
the EU H2020 Program under the scheme INFRAIA-01-2018-2019: Research and Innovation action grant agreement number 871042 SoBigData++: European Integrated Infrastructure for Social Mining and Big Data Analytics.
Fabrizio Falchi has been also supported by 
AI4Media - A European Excellence Centre for Media, Society and Democracy, H2020 ICT-48-2020, grant 95191.

%\nolinenumbers

% Either type in your references using
% \begin{thebibliography}{}
% Text
% \end{thebibliography}
%
% or
%
% Compile your BiBTeX database using our plos2015.bst
% style file and paste the contents of your .bbl file
% here. See http://journals.plos.org/plosone/s/latex for 

\begin{figure}[!h]
\label{S1}
\centering\includegraphics[width=0.9\textwidth]{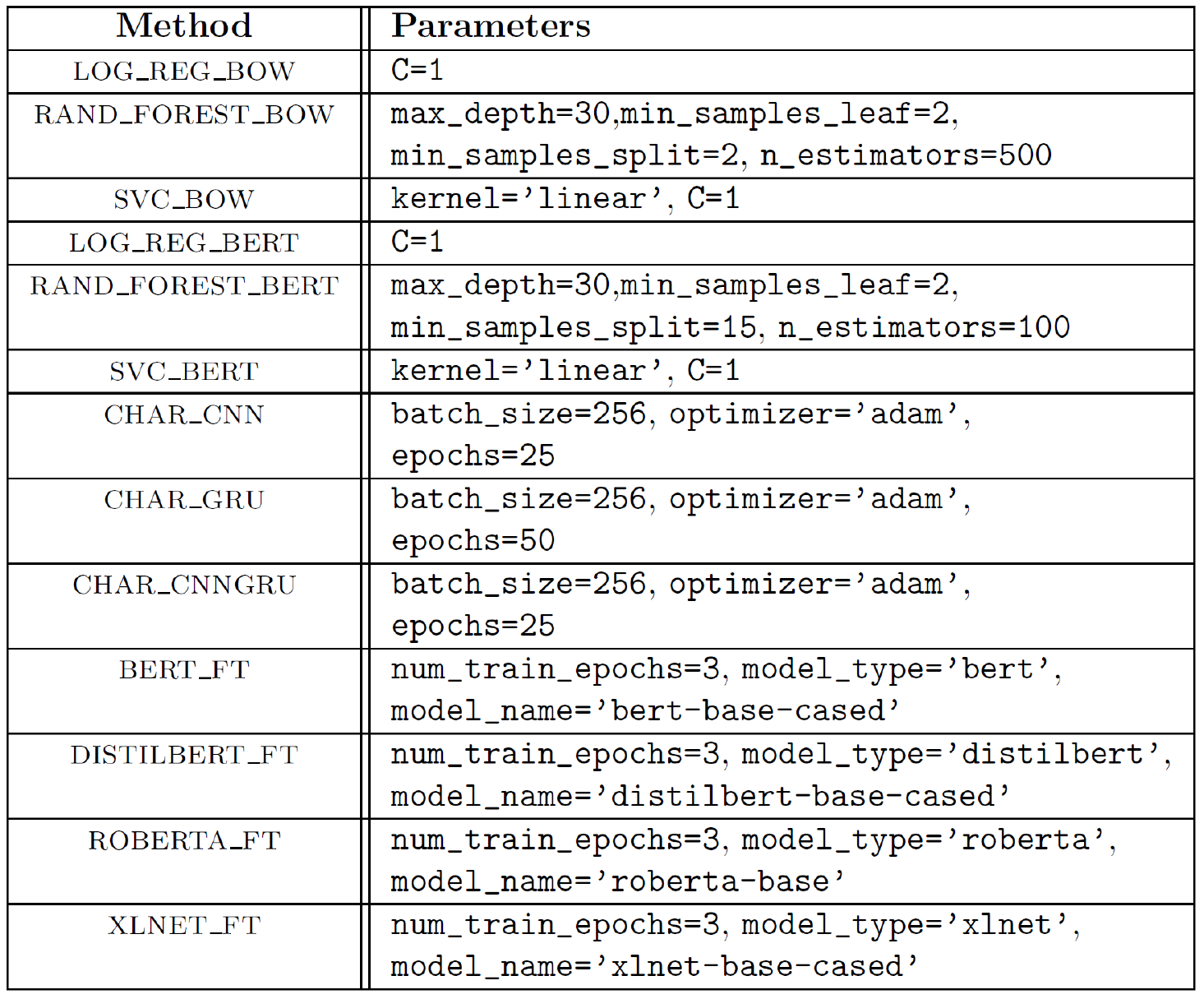}
\caption{{\bf Supp. info 1: Parameter values.} Parameter values used in the final experimentation on the test set.
}
\end{figure}

\begin{figure}[!h]
\centering\includegraphics[width=0.8\textwidth]{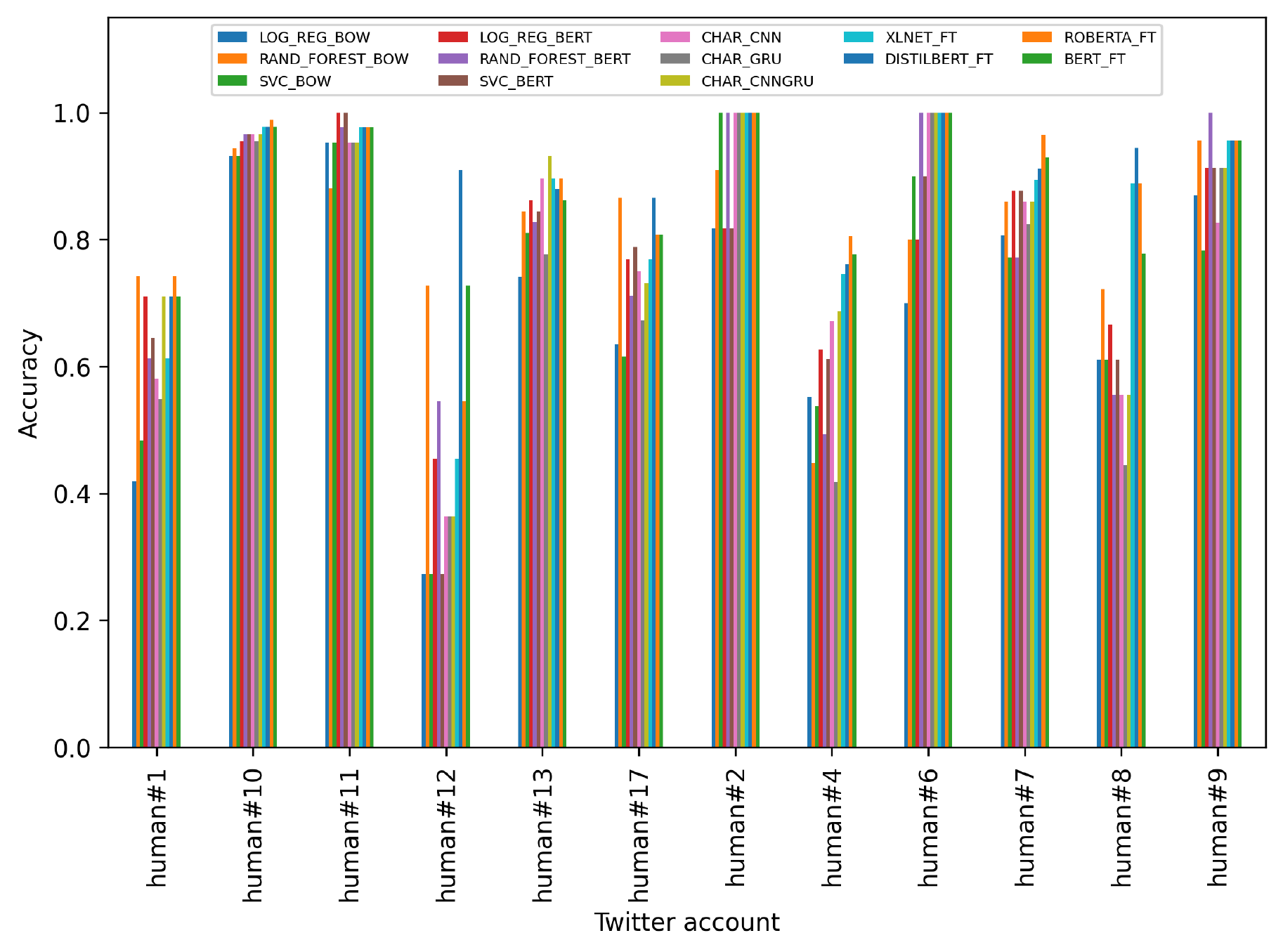}
\caption{{\bf Supp. info 2: Acc. over humans.} Detection Accuracy of tested methods on 'human' accounts with at least 10 examples.
}
\end{figure}

\begin{figure}[!h]
\centering\includegraphics[width=0.8\textwidth]{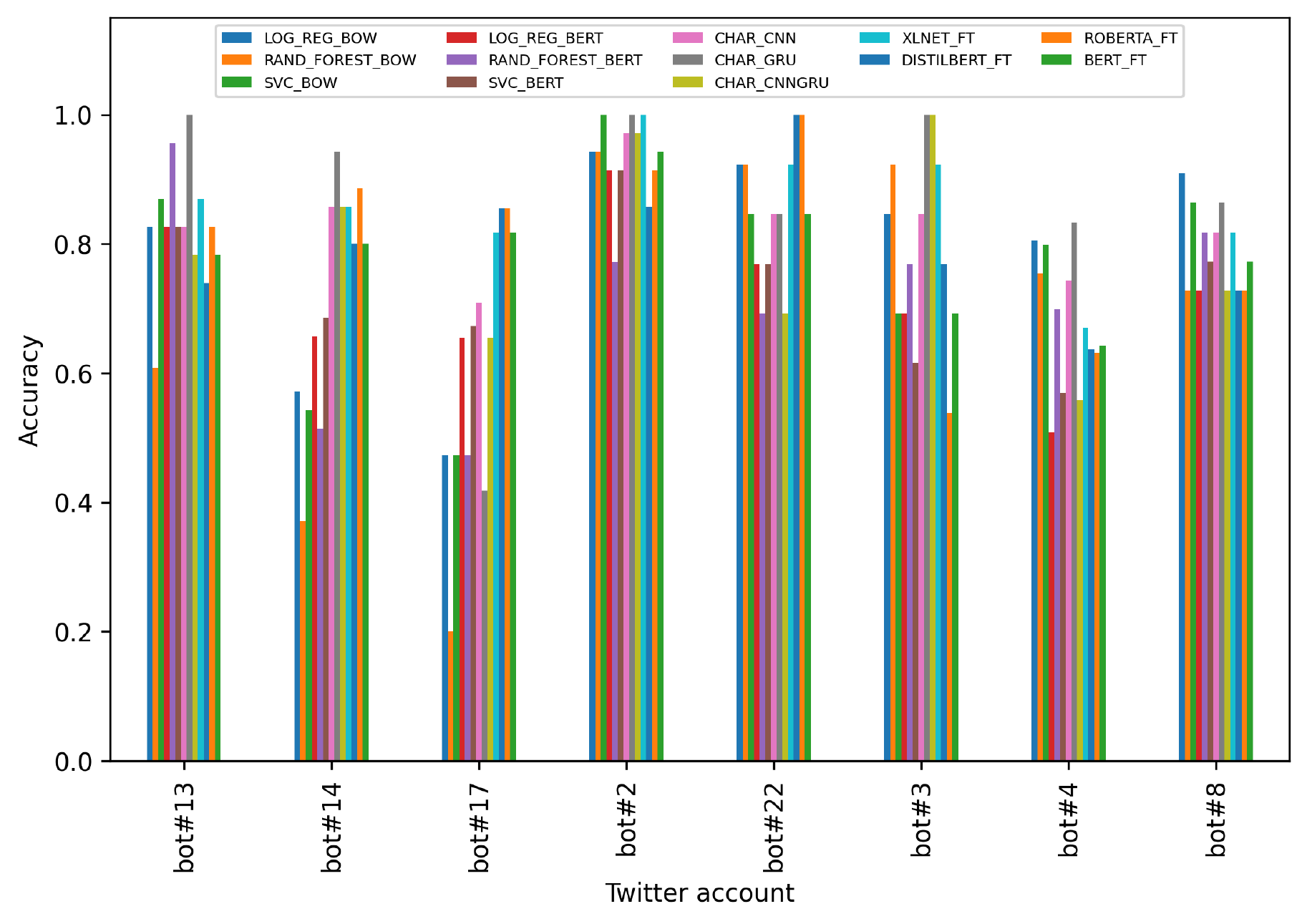}
\caption{{\bf Supp. info 3:  Acc. over GPT-2.} Detection Accuracy of tested methods on 'gpt2' accounts with at least 10 examples.
}
\end{figure}

\begin{figure}[!h]
\centering\includegraphics[width=0.8\textwidth]{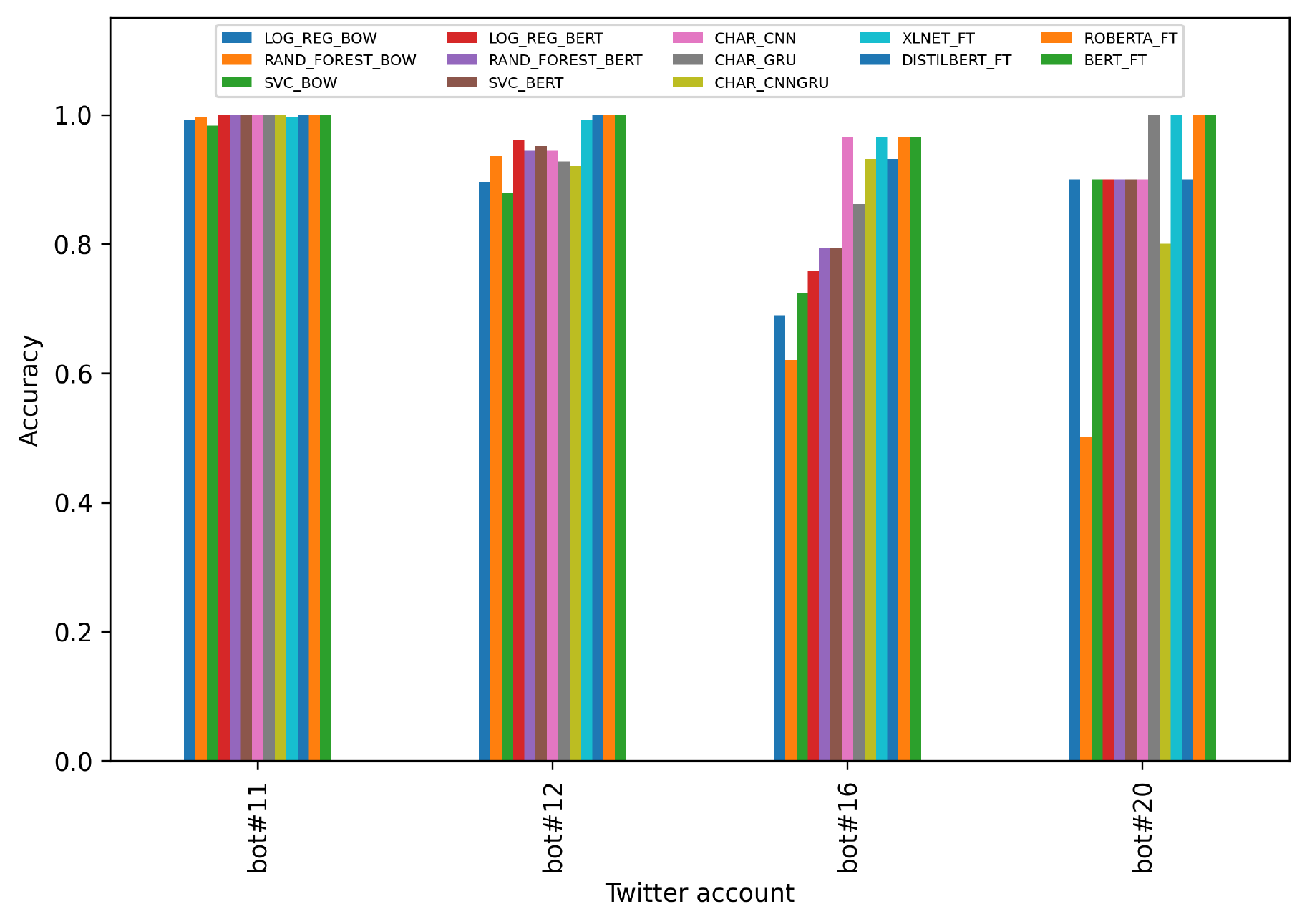}
\caption{{\bf Supp. info 4: Acc. over RNN.} Detection Accuracy of tested methods on 'rnn' accounts with at least 5 examples.}
\end{figure}

\begin{figure}[!h]
\centering\includegraphics[width=0.8\textwidth]{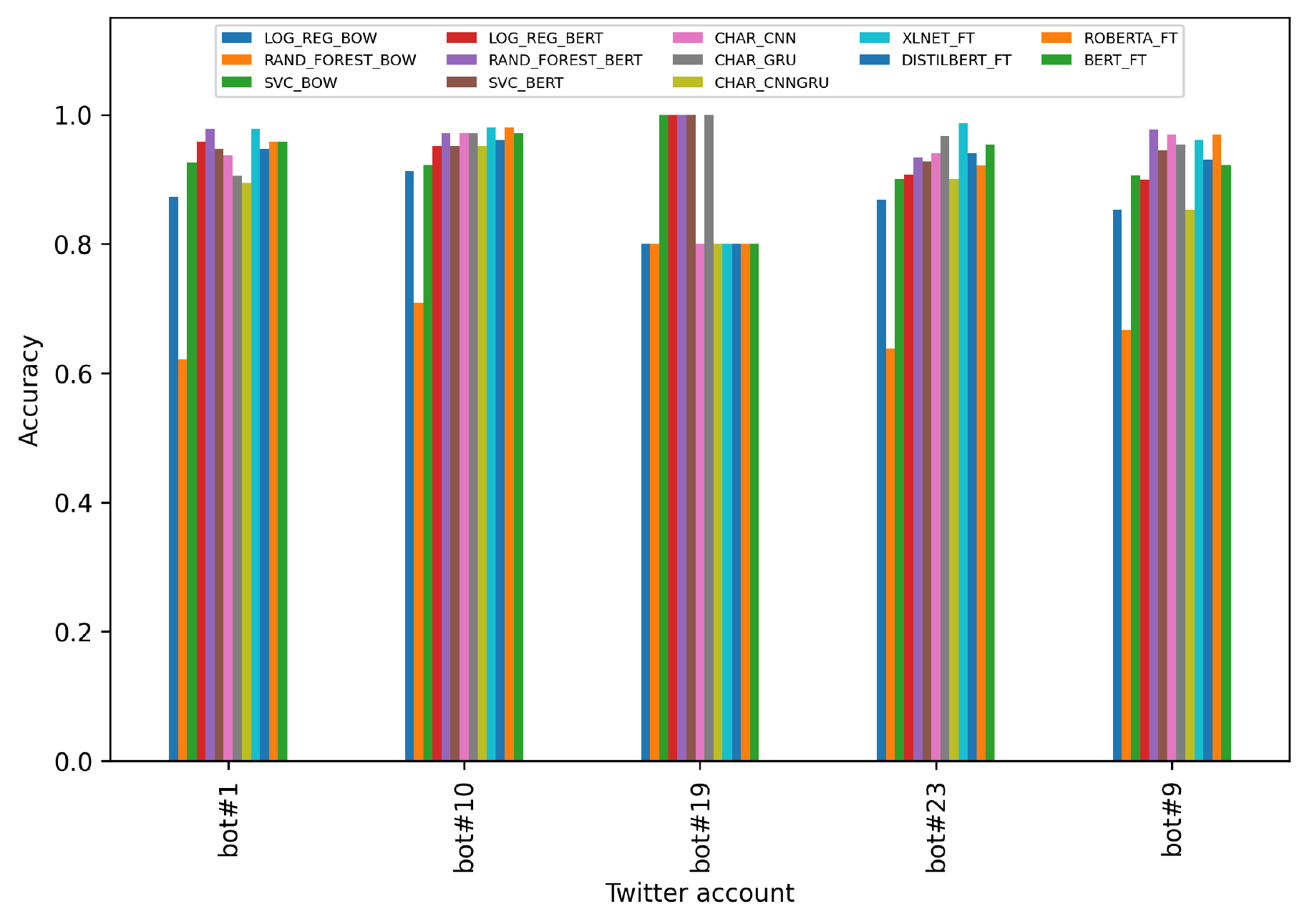}
\caption{{\bf Supp. info 5: Acc. over \emph{others}.} Detection Accuracy of tested methods on 'others' accounts with at least 5 examples.}
\end{figure}

\end{document}